\documentclass[10pt,twocolumn,letterpaper]{article}

\usepackage{iccv}
\usepackage{times}
\usepackage{epsfig}
\usepackage{graphicx}
\usepackage{amsmath}
\usepackage{amssymb}

\usepackage{multirow}
\usepackage{eucal}
\DeclareMathOperator*{\argmin}{argmin}
\DeclareMathOperator*{\argmax}{argmax}
\usepackage{enumitem}
\usepackage{diagbox}
\graphicspath{ {./images/} }
\usepackage{caption}
\newcommand{\zerodisplayskips}{%
  \setlength{\abovedisplayskip}{0pt}%
  \setlength{\belowdisplayskip}{0pt}%
  \setlength{\abovedisplayshortskip}{0pt}%
  \setlength{\belowdisplayshortskip}{0pt}}

\usepackage[pagebackref=true,breaklinks=true,letterpaper=true,colorlinks,bookmarks=false]{hyperref}

\iccvfinalcopy 

\begin{document}

\title{Confidence Calibration for Domain Generalization under Covariate Shift}

\author{Yunye Gong$^1$, Xiao Lin$^1$, Yi Yao$^1$, Thomas G. Dietterich$^2$, Ajay Divakaran$^1$, and Melinda Gervasio$^1$\\$^1$SRI International, $^2$School of Electrical Engineering and Computer Science, Oregon State University\\
\tt\small$^1${first.last@sri.com},$^2${tgd@oregonstate.edu}}

\maketitle
\begin{abstract}
 
   Existing calibration algorithms address the problem of covariate shift via unsupervised domain adaptation. However, these methods suffer from the following limitations: 1) they require unlabeled data from the target domain, which may not be available at the stage of calibration in real-world applications and 2) their performance depends heavily on the disparity between the distributions of the source and target domains. 
   To address these two limitations, we present novel calibration solutions via domain generalization. 
   Our core idea is to leverage multiple calibration domains to reduce the effective distribution disparity between the target and calibration domains for improved calibration transfer without needing any data from the target domain. We provide theoretical justification and empirical experimental results to demonstrate the effectiveness of our proposed algorithms. Compared against state-of-the-art calibration methods designed for domain adaptation, we observe a decrease of $8.86$ percentage points in expected calibration error or, equivalently, an increase of $35$ percentage points in improvement ratio for multi-class classification on the Office-Home dataset.

\end{abstract}

\vspace{-5mm}
\section{Introduction}
\begin{figure}[!t]
\begin{center}
\hbox{\includegraphics[scale=0.35]{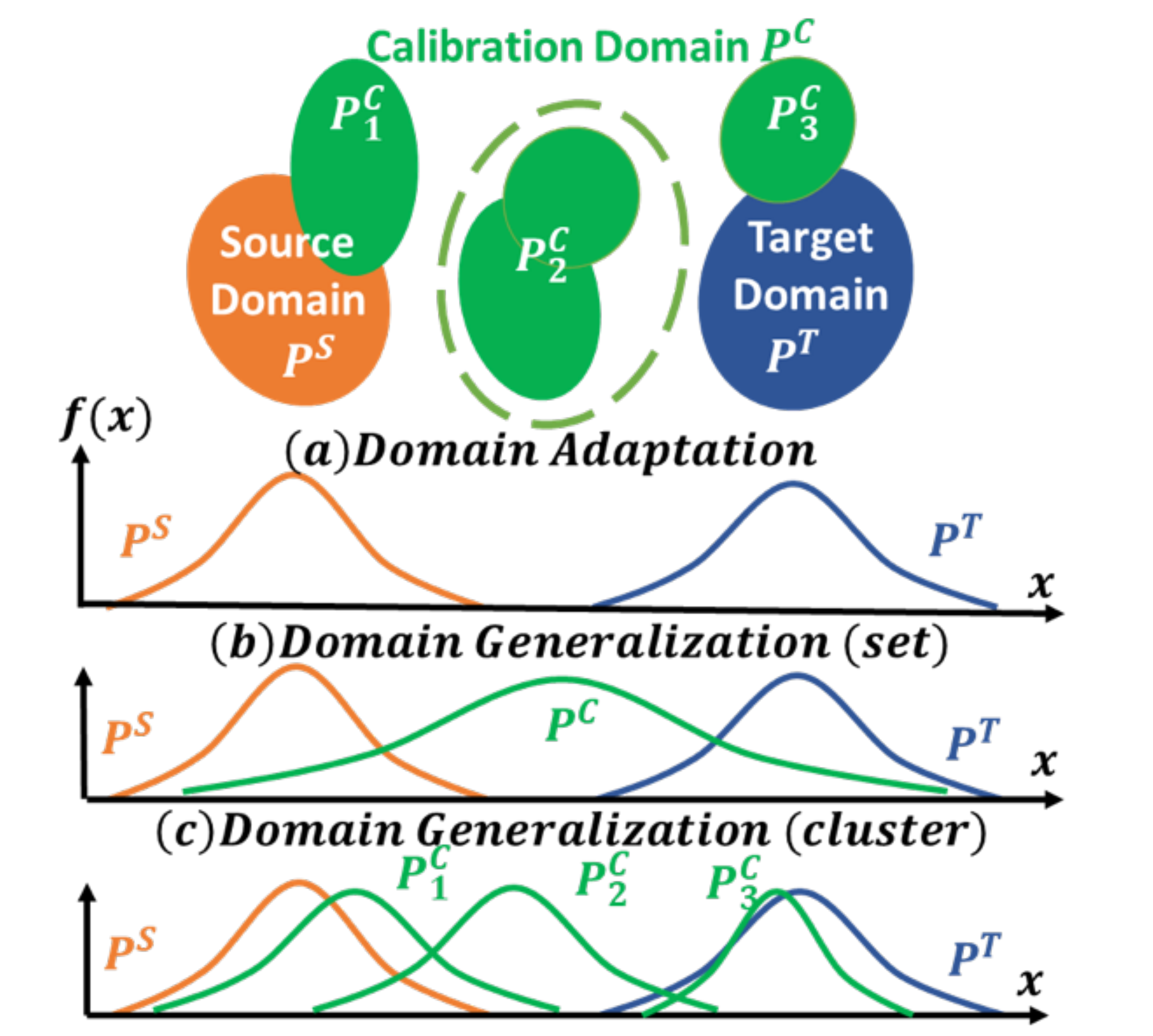}}
\end{center}
\vspace{-3em}
\caption{Calibration for domain adaptation with a single source domain may suffer from a large variance of the density ratio (i.e., $P^T/P^S$) caused by disjoint $P^S$ and $P^T$ and, therefore, a large calibration error as shown in (a). Our proposed calibration algorithms for domain generalization leverage multiple calibration domains to reduce the disparity between $P^C$ and $P^T$ for a decreased variance of the density ratio $P^T/P^C$ and, in turn, improved calibration performance, as shown in (b) and (c).} 
\label{fig:teaser}
\vspace{-2.5em}
\end{figure}
\vspace{-2mm}
Deep neural networks (DNNs) have demonstrated high accuracy for tasks such as classification and detection given adequate data and supervision~\cite{xie20,touvron19}. However, for real-world applications, the ability to indicate how much users should trust model predictions can be even more crucial than just having an accurate but unpredictable model~\cite{amodei16,bhavya19,toreini20}. While discriminative networks provide confidence scores that can be used as a heuristic measure of the probability of correct classification, such scores are not guaranteed to match the true probabilities of correct classification~\cite{guo17}. A recent development, referred to as model calibration, addresses this problem directly~\cite{platt99,guo17}. 

A classifier is calibrated with respect to a distribution (or a dataset sampled from that distribution) if its predicted probability of being correct matches its true probability. If the distribution changes, calibration is usually lost, and this has been demonstrated empirically~\cite{pampari20}. 
Recent work has begun to investigate the problem of calibration in the context of transfer learning, specifically in an unsupervised domain adaptation scenario under the assumption of covariate shift~\cite{park20,wang20,pampari20}. However, these methods need at least unlabeled data from the target domain, which may not be available at the stages of training and calibration in real-world applications. Furthermore, as these methods are designed to handle a single source domain, there may be an undesired disparity between selected source and target domains either due to the limited availability of sources or uncertainties (e.g., extreme weather/unexplored terrain) in the target. Fig.~\ref{fig:teaser}(a) notionally depicts such a scenario. This disparity results in a large variance of the density ratio defined as $P^T/P^S$, which significantly degrades the accuracy of calibration~\cite{pampari20,cortes10}.  

To tackle the aforementioned limitations of calibration transfer via domain adaptation, we focus instead on calibration for domain generalization.   
 Our key idea is to use multiple source domains and cluster their labeled data into groups. We then fit post-hoc calibration parameters to each group. The class probability of a test example is calibrated using the calibration parameters of the group that is nearest (in Euclidean distance) to the test example. By using many calibration domains, we increase the likelihood of overlap in the distributions, which in theory will improve the effectiveness of cross-domain calibration~\cite{pampari20,cortes10}. By learning calibration parameters separately for each group, we increase the likelihood that each test query will be adjusted by the best calibration correction. 

We study two calibration methods within each cluster. Both are based on temperature scaling~\cite{guo17}. The first computes a fixed scaling temperature for each group (Fig.~\ref{fig:teaser}(c)). The second fits a regression model to these fixed temperatures to enable extrapolating the temperature to points outside the clusters. We compare these two methods against a baseline that computes one temperature for scaling based on the union of all the calibration data (Fig.~\ref{fig:teaser}(b)). We refer to our methods as \emph{cluster-level} and the baseline as \emph{set-level}. Notably, while Fig.~\ref{fig:teaser} depicts a notional scenario where the source and target domains have no overlap and the calibration domains bridge the gap, our methods will also work well in cases where the source is closer to the target, as long as at least one of the calibration domains is also close to the target.

Our major contributions include the following:

1) We propose novel solutions to calibrate a classification model for domain generalization. Our proposed algorithms are trained to produce accurate confidence predictions without needing any data from the target domain.

2) We provide theoretical error bounds for our proposed calibration methods and demonstrate the advantage of our methods in maximizing the overlap between the supports of the target and calibration distributions, a critical factor that determines the generalization performance of calibration.

3) We justify the proposed algorithms with experimental results on real-world data. A decrease of $8.86$ percentage points in expected calibration error or, equivalently, an increase of $35$ percentage points in improvement ratio is achieved on the Office-Home dataset~\cite{officehome} compared against state-of-the-art (SOTA) calibration methods designed for domain adaptation. 

\vspace{-2mm}
\section{Related Work}\label{sec:relatedworks}
\vspace{-2mm}
\noindent\textbf{Calibration.}
Existing calibration methods provide post-hoc correction for classification models so that their confidence scores better match the true probabilities of correct classification~\cite{zad01,zad02,ECE,platt99,guo17}. Among those, Platt Scaling~\cite{platt99} provides a parametric solution for binary classification. It learns a logistic regression model with two scalar parameters that maps the initial predicted probabilities to calibrated probabilities. It is trained on a holdout validation set with respect to the negative log-likelihood (NLL) loss. Matrix scaling and Vector scaling~\cite{guo17} are two extensions of Platt scaling to multi-class classification problems, where a linear transformation is applied to the logit vectors before the softmax operation. Given a classification model, the additional linear layer is finetuned on the validation set with respect to NLL. In this case, classification accuracy is affected by calibration. Temperature scaling is another special case of Platt scaling. Here, a scalar temperature parameter is applied to scale the logit vectors without changing class predictions. The temperature is optimized on the validation set with respect to NLL and can be interpreted as the solution of a constrained entropy maximization. Alexandari et al.~\cite{alexandari20} investigated variants of vector and temperature scaling including no-bias vector scaling and bias-corrected temperature scaling in the context of domain adaptation under label shift. Our proposed algorithms are all based on temperature scaling~\cite{guo17}.

\noindent\textbf{Domain generalization vs.~domain adaptation.}
Transfer learning is generally challenging for deep learning, as models trained on one domain (source) can suffer performance drops when evaluated on test data from a different domain (target). One type of transfer learning is domain adaptation~\cite{wang18,csurka17}, which seeks to improve target domain performance by leveraging data from both source and target domains. Specifically, unsupervised domain adaptation (UDA)~\cite{wilson20} addresses the problem when only unlabeled data is available from the target domain. Multiple UDA methods have been developed based on strategies such as learning domain invariant features~\cite{long17,kang19,sun19,pan19} and learning mappings between domains~\cite{hoffman18,shriva17}. 

An alternative to domain adaptation is domain generalization, which aims at robust transfer without any data, either labeled or unlabeled, from the target domain at the training stage, by leveraging information from multiple related source domains. Ghifary et al.~\cite{ghifary15} propose a multi-task autoencoder (MTAE) that jointly reconstructs analogous views of a source image over multiple domains to acquire robust features for generalization in the context of object recognition. Li et al.~\cite{li18} minimize maximum mean discrepancy (MMD) to align distributions from different domains. Several recent studies adopt model-agnostic meta-learning (MAML) originally proposed for few-shot learning~\cite{finn17}. For instance, Li et al.~\cite{mldg} propose meta-learning for domain generalization (MLDG) using model-agnostic optimization across domains instead of across tasks. 
Balaji et al.~\cite{balaji18} apply meta-learning to learn a generalizable regularizer for the classification layers instead of the full network. Dou et al.~\cite{dou19} introduce complementary losses to encourage class alignment across domains and improve compactness of class-specific clusters.

All of this work is devoted to learning models to improve generalization with respect to \emph{classification accuracy}. In comparison, our proposed methods focus on calibrating a classifier using multiple related domains to improve confidence scores so that they are better calibrated in the unseen target domain (i.e., \emph{fidelity of confidence scores}).

\noindent\textbf{Calibration for domain adaptation.} Several recent papers investigate the problem of calibration in the context of transfer learning, specifically in an unsupervised domain adaptation scenario under the assumption of covariate shift~\cite{park20,wang20,pampari20}. These studies adopt similar frameworks based on estimating importance weights that describe the density ratio between the source and target distributions. The weights are estimated by learning a discriminator distinguishing source samples from target samples. The calibration loss in the target domain can then be formulated as a weighted version of the original loss in the source domain. Calibration loss is quantified using Brier Score~\cite{park20}, NLL~\cite{pampari20}, and expected calibration error (ECE)~\cite{wang18}. 

While these recent efforts are the most relevant to our work, we address an arguably more challenging problem. Instead of calibrating classifiers using unlabeled target data, we calibrate classifiers without any data, either labeled or unlabeled, from the target domain.
\vspace{-2mm}
\section{Background}
\label{sec:background}
\vspace{-2mm}
\noindent\textbf{Calibration.}
Let $x,y$ denote the data and label drawn from a joint distribution $P(x,y)$. Let $\phi(\cdot)$ be a learned multi($K$)-class classification model that projects each sample $x_i$ to a logit vector $z_i$ with $K$ dimensions. The class prediction $\hat{y_i}$ and confidence prediction $\hat{p_i}$ can be expressed as
\begin{equation}
    \hat{p_i}=\max_k \sigma(z_i)^{(k)}\quad\hat{y_i}=\argmax_k\sigma(z_i)^{(k)},
\end{equation} 
where $\sigma$ denotes the the softmax function:
\begin{equation}
\sigma(z^{(k)})=\frac{\exp(z^{(k)})}{\sum_{j=1}^K\exp(z^{(j)})}.
\end{equation}
Miscalibration refers to the problem where confidence predictions $\hat{p_i}$ do not match the true probabilities of correct classification.
The goal of calibration is to adjust the confidence so that $\mathbb{P}(\hat{y}=y|\hat{p}=p)=p, \forall p\in[0,1]$ \cite{guo17}.

\noindent\textbf{Temperature scaling~\cite{guo17}.}
A scalar $t>0$ is applied to adjust the confidence prediction:
\vspace{-0.5em}
\begin{equation}
    \hat{p_i}=\max_k\sigma(z_i/t)^{(k)}.
\end{equation}
The value of $t$ is optimized over a small validation set with respect to the same NLL loss used in classification training: 
\begin{equation}
    t^*=\argmin_{t}\mathbb{E}_{x,y\sim P(x,y)}\mathcal{L}(\phi(x),y,t),
\end{equation} 
where $\mathcal{L}$ denotes the NLL loss. Note that temperature scaling does not affect the overall classification accuracy, as the same $t$ is applied to all classes. 

\noindent\textbf{Expected calibration error (ECE)~\cite{ECE}.} To measure calibration accuracy, we employ the ECE metric. Given a set of class predictions and corresponding confidence predictions, ECE is computed by grouping test samples into $M$ bins of equal width based on confidence values. Let $B_m$ denote the set of indices where $B_m=\{i|\hat{p_i}\in(\frac{m-1}{M},\frac{m}{M}]\}$. The classification accuracy and average confidence for each bin are computed as
\vspace{-1em}
\begin{align}
    acc(B_m)&=\frac{1}{|B_m|}\sum_{i\in B_m}1(\hat{y_i}=y_i)\\
    conf(B_m)&=\frac{1}{|B_m|}\sum_{i\in B_m}\hat{p_i}.
\end{align}
 A well-calibrated model should reduce the mismatch between classification accuracy and confidence prediction. Therefore, ECE is computed as the weighted sum of the mismatch over bins:
 \vspace{-1mm}
\begin{align}
    ECE = \sum^{M}_{m=1}\frac{|B_m|}{N}|acc(B_m)-conf(B_m)|,
\end{align}
where $N$ is the total number of the samples.
In the case of $M=1$, ECE reduces to the absolute error between the average confidence prediction and the classification accuracy over the entire test set. 

\noindent\textbf{Calibration for domain adaptation.}
Let $P^{S}(x,y)$ and $P^{T}(x,y)$ denote the source and target distributions, respectively. Covariate shift between distributions refers to the assumption that $P^{T}(x)\neq P^{S}(x)$ while $P^{T}(y|x)=P^{S}(y|x)$.
Following a formulation similar to those in domain adaptation approaches (Theorem 4.1 in~\cite{pampari20}), the desired calibration loss 
can be expressed as
\vspace{-1mm}
\begin{align}
    &\mathbb{E}_{x,y\sim P^{T}(x,y)}\mathcal{L}(\phi(x),y,t) \nonumber\\
    &= \int_x\int_y\mathcal{L}(\phi(x),y,t)P^{T}(x,y)\,dx\,dy\nonumber\\
    &=\int_x\int_y\mathcal{L}(\phi(x),y,t)\frac{P^{T}(x)P^{T}(y|x)}{P^{S}(x)P^{S}(y|x)}P^{S}(x,y)\,dx\,dy\nonumber\\
    &=\mathbb{E}_{x,y\sim P^{S}(x,y)}w_S(x)\mathcal{L}(\phi(x),y,t)
    \label{Eq:transcal}
\end{align} 
for $\{x|P^{T}(x)>0\}\subseteq\{x|P^{S}(x)>0\}$, where the importance weight $w_S(x)=\frac{P^{T}(x)}{P^{S}(x)}$ is the density ratio.
\vspace{-2mm}
\section{Method}\label{sec:method}
\vspace{-2mm}
\noindent In contrast to confidence calibration under the single-source single-target unsupervised domain adaptation scenario, we consider domain generalization with multiple source domains that are related but different from the holdout target domain. In this case, we use $S$ (source) to denote the group of domains used for training the classifier, $C$ (calibration) to denote the group of domains used for calibrating the given classifier, and $T$ (target) to denote the group of holdout test domains that are completely unseen at both the classifier training and calibration stages. Accordingly, the desired calibration loss is given by
\vspace{-1.5mm}
\begin{align}
    &\mathbb{E}_{x,y\sim P^{T}(x,y)}\mathcal{L}(\phi(x),y,t) \nonumber\\
    &= \int_x\int_y\mathcal{L}(\phi(x),y,t)P^{T}(x,y)\,dx\,dy\nonumber\\
    &=\int_x\int_y\mathcal{L}(\phi(x),y,t)\frac{P^{T}(x)P^{T}(y|x)}{P^{C}(x)P^{C}(y|x)}P^{C}(x,y)\,dx\,dy\nonumber\\
    &=\mathbb{E}_{x,y\sim P^{C}(x,y)}w_C(x)\mathcal{L}(\phi(x),y,t),
    \label{Eq:transcal_set}
\end{align} 
for $\{x|P^{T}(x)>0\}\subseteq\{x|P^{C}(x>0\}$, where $w_C(x)=\frac{P^T(x)}{P^C(x)}$ denotes the density ratio between the target and calibration domains. 

Following~\cite{pampari20,wang20}, we derive the gap between the calibration loss and the oracle loss using the true target distribution $P^T(x,y)$ to show that the variance of the density ratio $w_C(x)$ or equivalently the divergence between $P^T(x)$ and $P^C(x)$ is critical for calibration transfer. The same observations apply to calibration for domain adaptation, where the variance of $w_S(x)$ or the divergence between $P^T(x)$ and $P^S(x)$ is critical. For simplicity, we use $w(x)$ to denote $w_C(x)$ or $w_S(x)$ when these two are interchangeable. The gap is given by
\vspace{-1.5mm}
\begin{align}
    &\big|\mathbb{E}_{ P^{C}(x,y)}\mathcal{L}(\phi(x),y,t) - \mathbb{E}_{ P^{T}(x,y)}\mathcal{L}(\phi(x),y,t)\big|\nonumber\\
    =&\big|\int_x\int_y(1-w_C(x))\mathcal{L}(\phi(x),y,t)P^{C}(x,y)\,dx\,dy\big|\nonumber\\
    =&\big|\mathbb{E}_{ P^{C}(x,y)}\big[(1-w_C(x))\mathcal{L}(\phi(x),y,t)\big]\big|\\
    \leq&\sqrt{\mathbb{E}_{ P^{C}(x)}\big[(1-w_C(x))^2\big]\mathbb{E}_{ P^{C}(x,y)}\big[\mathcal{L}(\phi(x),y,t)^2\big]}\label{Eq:CS}\\
    \leq&\frac{1}{2}(\mathbb{E}_{P^{C}(x)}\big[(1-w_C(x))^2\big]+\mathbb{E}_{P^{C}(x,y)}\big[\mathcal{L}(\phi(x),y,t)^2\big]),
    \label{Eq:AMGM}
\end{align}\par
\vspace{-2mm}
where the inequality in Eq.~\ref{Eq:CS} follows from the Cauchy-Schwarz Inequality and the inequality in Eq.~\ref{Eq:AMGM} follows from the inequality of arithmetic and geometric means. This formulation can also be interpreted as the bound of the bias of the estimated loss given
\vspace{-1.5mm}
\begin{equation}
    \hat{w_C}(x)=\mathbb{E}_{P^C}[w_C(x)]=1
\end{equation}
as an estimator of $w_C(x)$. We exploit this property to design calibration algorithms bypassing the direct computation of $w_C(x)$ due to the lack of target data at the classifier training and calibration stages. 

Given a fixed classification model $\phi$, the second term in Eq.~\ref{Eq:AMGM} is computed based on calibration data. Therefore, only the first term is affected by the shift between the calibration and target domains. Following Cortes et al.,~\cite{cortes10}, the first term in Eq.~\ref{Eq:AMGM} can be expressed as 
\begin{align}
    \mathbb{E}_{P^{C}}\big[(w_C(x)-1)^2\big]&=\mathbb{E}_{P^{C}}\big[(w_C(x)-\mathbb{E}_{P^C}[w_C(x)])^2\big]\nonumber\\&=\text{Var}(w_C(x))\nonumber\\
    &=d_2\big(P^{T}(x)||P^{C}(x)\big)-1,
    \label{Eq:var}
\end{align}
where $d_{\alpha}(P||Q)=\big[\sum_x\frac{P^{\alpha}(x)}{Q^{\alpha-1}(x)}\big]^{\frac{1}{\alpha-1}}$ with $\alpha>0$ is the exponential in base 2 of the Renyi-divergence~\cite{renyi60} between distributions $P$ and $Q$.

The calibration errors are dominated by the variance of $w_C(x)$ given by $\text{Var}(w_C(x))=d_2\big(P^{T}(x)||P^{C}(x)\big)-1$. Similarly, for domain adaptation, we have $\text{Var}(w_S(x))=d_2\big(P^{T}(x)||P^{S}(x)\big)-1$. Intuitively we seek to reduce the variance of $w(x)$, or equivalently, the divergence between target and source distributions for domain adaptation or target and calibration distributions for domain generalization. 
If there exist large shifts between the source and target domains, the density ratio is unbounded over $\{x|P^T(x)\neq0,P^S(x)=0\}$ leading to large variance of the density ratio. 
Let $\{x|P(x)>0\}$ be the support of a distribution $P(x)$. Reducing the variance of $w(x)$  requires larger overlap between the supports of $P^T(x)$ and $P^S(x)$ for domain adaptation or between $P^T(x)$ and $P^C(x)$ for domain generalization. 

For domain adaptation with fixed source and target domains~\cite{park20,wang20,pampari20}, there is limited room for calibration to adjust such overlap. In contrast, for domain generalization, we can manipulate the learning of calibration models over different calibration domains to maximize such overlap. Motivated by these theoretical advantages of using multiple sources, we propose calibration algorithms for domain generalization (Fig.~\ref{fig:block_diagram}) including set- (Sec. \ref{sec:set}) and cluster-level approaches (Sec. \ref{sec:cluster}). Note that while maximizing the overlap between $P^T(x)$ and $P^S(x)$ (or $P^C(x)$) can be achieved via feature alignment, commonly used for domain adaptation, that is out of the scope of this paper. 

\vspace{-1mm}
\subsection{Set-level calibration}
\label{sec:set}
\vspace{-2mm}
 
\begin{figure}[!t]
\centering
\hbox{\includegraphics[scale=0.3]{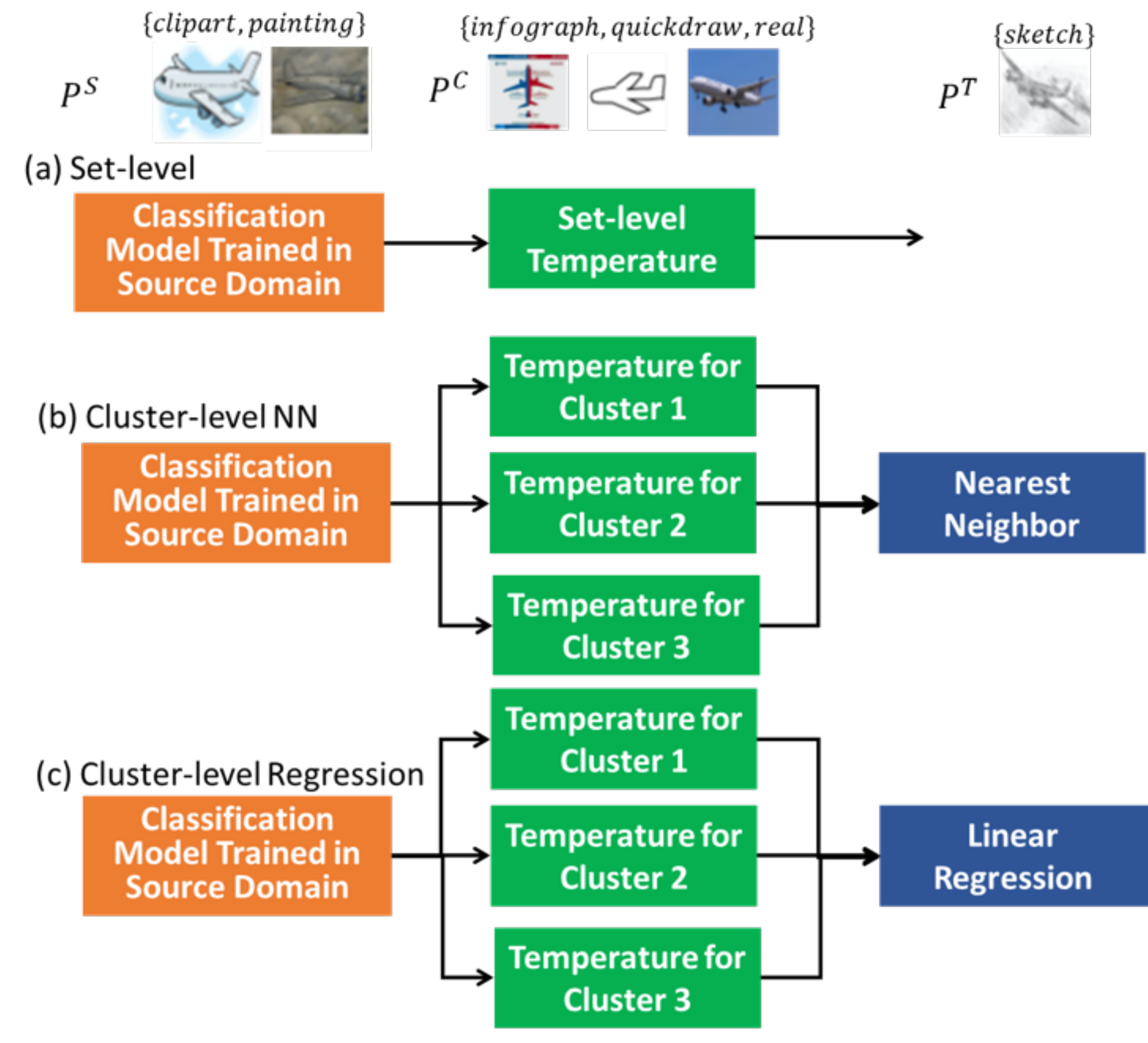}}
\vspace{-1em}
\caption{Block diagram of proposed calibration algorithms for domain generalization: (a) set-level, (b) cluster-level NN, and (c) cluster-level regression.}
\label{fig:block_diagram}
\vspace{-2em}
\end{figure}

 Set-level calibration is our baseline method. We learn the temperature $t$ using multiple calibration domains $C$. Only a small set of data from $C$ is required, and temperature scaling is applied post-hoc to the classification model $\phi$ trained on the source domain $S$. The temperature is learned via
 \begin{align}
     t^* = \argmin_t\mathbb{E}_{x,y\sim P^{C}(x,y)}\mathcal{L}(\phi(x),y,t),
 \end{align} 
where $P^C(x,y)$ is the joint distribution over all calibration domains, meaning that each calibration domain is treated equally. We refer to this algorithm as set-level calibration (Fig. \ref{fig:block_diagram} (a)), since a single temperature is learned with respect to all calibration data and applied to all test data. 

By leveraging multiple related domains, $P^C$ is more likely to be better aligned with $P^T$ especially for scenarios where $P^S$ and $P^T$ are distant as shown in Fig.~\ref{fig:teaser}(b). This leads to a density ratio $w_C(x)$ with fewer unbounded values and, thus, a smaller variance. As a result, better calibration transfer can be achieved.

\vspace{-1mm}
 \subsection{Cluster-level calibration}
 \label{sec:cluster}
 \vspace{-1mm}
 Learning the temperature at the set level assumes the same optimal scaling for all samples. Considering calibration data from multiple domains, it would be natural to relax this constraint so that different samples can have different optimal scaling. Several classic calibration algorithms~\cite{ECE, zad01,zad02} perform calibration based on binning the data according to their uncalibrated confidence scores. Motivated by these successes, we propose to group the calibration samples by the similarity of image features to correlate the optimal temperature scaling to feature distributions. 
 Then, during testing we can predict the most appropriate temperature given only a single test image feature from an unknown distribution.
 Specifically, 
 we perform $K$-means clustering~\cite{kmeans} on the image features (at the penultimate layer of ResNet18 feature extractors) using calibration data. The centroids of clusters are determined by minimizing the within-cluster sum of squares. For each cluster, we perform a standard temperature scaling. At the testing stage, we exploit two alternative methods to determine the most appropriate temperature for each test sample.
 
 \noindent\textbf{Nearest neighbor (NN).} In the first method, we simply assign a given sample from the test domain to the cluster whose centroid is the closest to the sample feature in Euclidean distance. Then we apply the corresponding optimal temperature for that cluster to calibrate the test sample (Fig.~\ref{fig:block_diagram} (b)). Intuitively, this procedure facilitates the alignment of the calibration and target distributions at the cluster level. Let $N$ denote the number of clusters, $P^C_j$ denote the distribution of calibration domain features that are grouped into cluster $j$, and  $P^T_j$ denote the unknown distribution of all test domain features that are assigned to the cluster $j$. The optimal temperature for each cluster $j$ is learned with respect to the following objective:
 \begin{equation}
     t^*_j=\argmin_t\mathbb{E}_{P^{C}_j}\big[\mathcal{L}(\phi(x),y,t)\big].
     \zerodisplayskips
 \end{equation} 
 Following a similar formulation in Eq.~\ref{Eq:transcal_set}-Eq.~\ref{Eq:var}, the oracle objective for test samples from $P^T_j$ can be expressed as
 \begin{equation}
     \mathbb{E}_{P^{T}_j}\big[\mathcal{L}(\phi(x),y,t)\big]=\mathbb{E}_{P^{C}_j}\big[w_{C,j}(x,y)\mathcal{L}(\phi(x),y,t)\big],
     \label{Eq:transcal_cluster}
     \zerodisplayskips
 \end{equation}
 for $\{x|P^{T}_j(x)>0\}\subseteq\{x|P^{C}_j(x)>0\}$. Here, $w_{C,j}(x)=\frac{P^T_j(x)}{P^C_j(x)}$, and the generalization of learned temperatures depends on the divergence $d_2\big(P^T_j(x)||P^C_j(x)\big)$. Since $P^C_j$ is chosen as the closest cluster to the test samples, chances are high that $P^T_j$ and $P^C_j$ are better aligned than the set-level distributions $P^T$ and $P^C$, resulting in a density ratio $w_{C,j}$ with a smaller variance (e.g., Fig.~\ref{fig:teaser}(c) with $j=3$). Therefore, cluster-level calibration holds the promise of further improved calibration transfer.  

 \noindent\textbf{Regression-based prediction.}
 Nearest neighbor can be considered as a special case of linear regression where the $j^{th}$ weight is set to $1$ while the others are set to $0$. We further investigate cluster-level calibration using learned weights as a more generalized scheme (Fig.~\ref{fig:block_diagram} (c)). 
 Specifically, we train a regression model that maps the mean feature of each cluster to its corresponding cluster-level optimal temperature. We can thus apply the learned regression model to any test feature to predict a proper temperature specific to the test instance. Let $R_\theta$ denote the regression model $R$ parameterized by $\theta$, $R_\theta$ is determined by minimizing the following mean-squared error:
 \begin{equation}
     \theta^* = \argmin_{\theta}\frac{1}{N}\sum_{j=1}^N\big( R_{\theta}(\mathbb{E}_{P^C_j}(x))-t^*_j\big)^2. 
           \zerodisplayskips
 \end{equation}
Essentially, we learn a function capturing the underlying mapping from features to the proper temperature for calibration and transfer it to the unknown target domain, instead of directly transferring the temperatures learned on calibration domains. 

\vspace{-2mm}
\section{Experiments}\label{sec:results}
\vspace{-2mm}
\subsection{Datasets}\label{sec:dataset}
\vspace{-2mm}

\noindent\textbf{Office-Home~\cite{officehome}} contains images of 65 classes across four domains corresponding to different rendering styles: Clipart (4365 images), Art (2427 images), Product (4439 images), and Real (4357 images). We split these four domains into three subsets: one domain as the source for training the classifier, two domains for post-hoc calibration of the classifier, and one holdout domain as the target for evaluating the calibrated classifier. We perform experiments for all 12 possible splits of domains, including the combinations where the source is relatively similar to the target judging from the image realism of the domains (e.g., Art as source, Clipart as target, Product and Real as calibration) and combinations where the source is relatively distant to the target (e.g., Clipart as source, Real as target, Art and Product as calibration). We randomly divide data from each domain into a \emph{Large} subset ($80\%$) and a \emph{Small} subset ($20\%$). We use the \emph{Large} subset for either training the classifier or evaluating the calibration performance and we use the \emph{Small} subset for either tuning the hyperparameters of classification training or calibrating the classifier. For each source domain, we train a ResNet18~\cite{resnet} initialized with parameters pre-trained on ILSVRC-1000. We extract image features at the penultimate layer of the network for clustering. For each domain split, we perform 1000 evaluations each with 1500 randomly selected samples from the target domain to estimate confidence intervals.

\begin{table*}[!ht]
\centering
\small
\begin{tabular}{l|rrrr|r}
\hline
Methods
& Clipart & Art & Product & Real & \textbf{Average} \\
\hline
Uncalibrated & 14.74$\pm$2.23 & 9.31$\pm$2.33& 5.66$\pm$1.23 & 4.92$\pm$0.94& 8.66$\pm$4.38 \\
Source-only & 18.02$\pm$1.81& 10.79$\pm$4.46& 6.90$\pm$2.78& 6.09$\pm$1.05& 10.45$\pm$5.50\\
Target-only (oracle) & 4.10$\pm$0.72& 3.56$\pm$0.59& 4.10$\pm$1.10& 4.01$\pm$0.68& 3.94$\pm$0.83\\
\hline
TransCal~\cite{wang20} & 21.37 & 20.60 & 11.37 & 8.23 & 15.39\\
WTS~\cite{pampari20} & 18.97 & 8.60 & 3.90 & 7.63& 9.78 \\
\hline
Set-level & 9.96$\pm$1.78 & 4.21$\pm$0.59& 5.57$\pm$2.12& 7.47$\pm$3.00& 6.80$\pm$2.99\\
Cluster-level NN & 11.43$\pm$1.83& 4.90$\pm$1.65& 5.10$\pm$1.36& 6.49$\pm$2.06& 6.98$\pm$3.17\\
Cluster-level Regression & 12.00$\pm$1.41 & 4.61$\pm$0.80& 5.11$\pm$1.12& 6.03$\pm$1.82& 6.94$\pm$3.26\\
\hline
Ensemble & 11.31$\pm$1.75 & 4.14$\pm$0.94 & 4.69$\pm$1.23 & 5.98$\pm$1.82 & 6.53$\pm$3.20\\
\hline
\end{tabular}
\vspace{-1em}
\caption{Calibration performance (ECE \%) on Office-Home averaged by target domain. 
}
\label{tab:officehome-ece-average}
\vspace{-1mm}
\end{table*}

\begin{table*}[!ht]
\centering
\small
\setlength\tabcolsep{3pt}
\begin{tabular}{r|rrr|rr|rrr|r}
\hline
& Uncalibrated & Source-only & \shortstack{Target-only\\(oracle)} & \shortstack{TransCal\\~\cite{wang20}} & \shortstack{WTS\\~\cite{pampari20}} & Set-level & \shortstack{Cluster\\-level NN} & \shortstack{Cluster-level\\Regression} & \textbf{Ensemble}\\
\hline
A$\rightarrow$ \text{C} & 11.84$\pm$0.76 & 16.95$\pm$0.77& 4.30$\pm$0.70 & 22.9& 12.8 & 10.98$\pm$0.76 & 12.54$\pm$0.81 & 13.10$\pm$0.81 & 12.53$\pm$0.76\\
P$\rightarrow$ \text{C} & 15.81$\pm$0.82& 20.30$\pm$0.84& 4.24$\pm$0.68& 40.4& 26.8 & 7.71$\pm$0.76 & 9.12$\pm$0.84 & 10.43$\pm$0.84 & 9.10$\pm$0.79\\
R$\rightarrow$ \text{C} & 16.58$\pm$0.86& 16.82$\pm$0.86& 3.76$\pm$0.66& 4.5& 17.3& 11.19$\pm$0.84 & 12.64$\pm$0.84 &12.48$\pm$0.83 & 12.28$\pm$0.84\\
\hline
C$\rightarrow$ \text{A} & 7.61$\pm$0.53 & 7.37$\pm$0.52 & 4.08$\pm$0.48 & 21.7 & 6.9 & 4.50$\pm$0.45 &4.72$\pm$0.48 &4.48$\pm$0.48 & 5.02$\pm$0.47\\
P$\rightarrow$ \text{A} & 12.52$\pm$0.53 & 17.05$\pm$0.53 & 3.43$\pm$0.46 & 18.5& 8.5 &3.68$\pm$0.46& 6.92$\pm$0.53 & 5.46$\pm$0.50 & 4.36$\pm$0.47\\
R$\rightarrow$ \text{A} & 7.80$\pm$0.48& 7.96$\pm$0.48& 3.16$\pm$0.41& 21.6 & 10.4& 4.44$\pm$0.46& 3.05$\pm$0.43 & 3.90$\pm$0.42 & 3.04$\pm$0.39 \\
\hline
C$\rightarrow$ \text{P} & 5.78$\pm$0.78& 5.57$\pm$0.76& 3.32$\pm$0.61& 14 & 6.4& 3.22$\pm$0.59& 3.50$\pm$0.59 & 3.95$\pm$0.63 & 3.25$\pm$0.57\\
A$\rightarrow$ \text{P} & 6.81$\pm$0.77 & 10.64$\pm$0.82& 5.35$\pm$0.71&9.3& 1.5&5.39$\pm$0.78& 6.26$\pm$0.79&6.08$\pm$0.76 & 5.56$\pm$0.75\\
R$\rightarrow$ \text{P} & 4.38$\pm$0.62 & 4.50$\pm$0.64& 3.63$\pm$0.59&15.6&3.8& 8.1$\pm$0.74& 5.54$\pm$0.68 & 5.30$\pm$0.67 & 5.27$\pm$0.68\\
\hline
C$\rightarrow$ \text{R} & 5.86$\pm$0.69 & 5.75$\pm$0.68& 3.67$\pm$0.62&6.4&5.7& 3.73$\pm$0.63& 4.00$\pm$0.63&3.95$\pm$0.64 & 3.82$\pm$0.66\\
A$\rightarrow$ \text{R} & 4.31$\pm$0.63 & 5.34$\pm$0.70& 4.19$\pm$0.62&5.1&6.4& 7.86$\pm$0.79& 6.75$\pm$0.73& 6.08$\pm$0.71 & 6.25$\pm$0.74\\
P$\rightarrow$ \text{R} & 4.59$\pm$0.64 & 7.18$\pm$0.78& 4.18$\pm$0.67&13.9&10.8&10.83$\pm$0.77&8.72$\pm$0.76& 8.05$\pm$0.78 & 7.88$\pm$0.75\\
\hline
\end{tabular}
\vspace{-1em}
\caption{Calibration performance (ECE \%) on  Office-Home. 
}
\label{tab:officehome-ece-split}
\vspace{-1mm}
\end{table*}

\noindent\textbf{DomainNet~\cite{DomainNet}} contains images of 345 classes across six domains corresponding to different rendering styles: Quickdraw (172500 images), Infograph (51605 images), Sketch (69128 images), Clipart (48129 images), Painting (72266 images),  and Real (172947 images). We split these six domains into three subsets: two domains as the source, three domains for the calibration, and one holdout domain as the target. We perform experiments for all 60 possible splits of domains, including the combinations where the source is relatively similar to the target judging from the image realism of the domains (e.g., Quickdraw and Sketch as source, Infograph as target, Clipart, Painting and Real as calibration) and combinations where the source is relatively distant to the target (e.g., Quickdraw and Sketch as source, Real as target, Clipart, Painting and Infograph as calibration). Following the train/test splits from~\cite{DomainNet}, we use the train split for training the classifier, a \emph{Small} subset ($10\%$) of the test split for calibration, and a \emph{Large} subset ($90\%$) of the test split for evaluation. We use a ResNet18 pretrained on ILSVRC-1000 as the feature extractor and train an MLP classifier. For each domain split, we conduct 1000 evaluations each with 10000 randomly selected samples from the target domain to estimate confidence intervals. 

\vspace{-1mm}
\subsection{Experimental settings}
\vspace{-1mm}
\noindent\textbf{Source-only calibration.}
We split the source domain as described in Sec~\ref{sec:dataset}, using the \emph{Large} subset to learn the classifier and the \emph{Small} subset to calibrate it. We directly evaluate the calibrated model on the holdout target domain. 
This experiment serves as a reference without calibration transfer.

\noindent\textbf{Target-only (oracle) calibration.}
Given a classifier trained on the source domain, we calibrate it using the \emph{Small} subset of the target domain data and evaluate the calibration on the \emph{Large} subset. This is an oracle experiment, since it uses the ground-truth labels and data from the target domain which are not available for the domain generalization setting. 
Consequently, this experiment sets the target performance for calibration transfer.

\noindent\textbf{Cross-domain calibration.}
Given a classifier trained on the source domain, we calibrate it via our algorithms described in Sec.~\ref{sec:method} using the \emph{Small} subsets from the calibration domains. We also average the logit outputs from these three methods as an additional ensemble-based approach. We evaluate the calibrated model on the \emph{Large} subset of the target domain which is unseen at both the classifier training and calibration stages. For cluster-level calibration, we use eight clusters for Office-Home and nine clusters for DomainNet. Our experiments suggest that the numbers of clusters have minor effects on calibration performance. For the cluster-level regression method, we choose a linear regression model considering the high dimensional feature space and availability of samples from calibration domains.

\begin{table*}[!ht]
\centering
\small
\begin{tabular}{l|rrrrrr|r}
\hline
Methods

& Quickdraw  & Infograph & Sketch & Clipart & Painting & Real & \textbf{Average} \\
\hline
Uncalibrated & 21.42$\pm$3.33 & 23.99$\pm$3.94 & 16.65$\pm$2.06 & 11.63$\pm$2.68 & 16.43$\pm$3.91 & 11.82$\pm$2.09 & 16.99$\pm$5.51\\
Source-only & 20.24$\pm$2.57& 24.58$\pm$4.52& 17.06$\pm$2.18& 11.67$\pm$3.26& 16.83$\pm$3.71& 11.24$\pm$3.21& 16.93$\pm$5.72 \\
Target-only (oracle) & 0.68$\pm$0.28 & 1.81$\pm$0.33 & 2.01$\pm$0.86 & 2.51$\pm$0.58 & 2.68$\pm$0.72 & 2.33$\pm$0.37& 2.00$\pm$0.87 \\
\hline
Set-level & 10.01$\pm$2.25 & 7.39$\pm$3.30 & 3.52$\pm$2.58 & 6.83$\pm$4.68& 5.87$\pm$4.43& 13.38$\pm$4.37& 7.83$\pm$4.87 \\
Cluster-level NN & 8.10$\pm$1.95& 7.91$\pm$2.24& 3.04$\pm$1.31& 6.06$\pm$2.74& 4.17$\pm$1.77& 9.35$\pm$2.65& 6.44$\pm$3.12\\
Cluster-level Regr. & 11.72$\pm$5.81& 11.93$\pm$6.74& 7.29$\pm$5.19& 8.49$\pm$3.74& 7.08$\pm$5.51& 9.55$\pm$5.39& 9.34$\pm$5.80\\
\hline
Ensemble & 9.81$\pm$2.54 & 9.51$\pm$2.99 & 3.12$\pm$1.59 & 5.71$\pm$2.50 & 3.66$\pm$2.54 & 8.15$\pm$2.74 & 6.66$\pm$3.67\\
\hline
\end{tabular}
\vspace{-1em}
\centering
\caption{Calibration performance (ECE \%) on DomainNet averaged for each target domain. 
\vspace{-1em}
}
\label{tab:DomainNet-ece}
\end{table*}
\vspace{-2mm}
\subsection{Results and Discussions} 
\vspace{-2mm}
Experimental results are summarized in Tables~\ref{tab:officehome-ece-average}-\ref{tab:DomainNet-ece}.
We report the mean and standard deviation of ECE scores (\%). Each column of Table~\ref{tab:officehome-ece-average} (for Office-Home) and Table~\ref{tab:DomainNet-ece} (for DomainNet)
lists the ECE scores for a specific target domain averaged over different domain splits across source and calibration domains. 
The last column lists the ECE scores averaged over different target domains. Table~\ref{tab:officehome-ece-split} lists the performance on Office-Home for each domain split (i.e., one combination of the source and target domains). 
ECE scores for the domain adaptation baselines~\cite{pampari20,wang20} are based on the reported results (using the CDAN~\cite{long18} method) from the original papers.
 
 The standard deviation of ECE, $\sigma_{ECE}$, is mainly affected by two factors: sample variations within a domain split and domain variations. Each $\sigma_{ECE}$ in Table~\ref{tab:officehome-ece-split} indicates the effect of sample variations within a fixed domain split. For all of the tested domain splits, we observe that $\sigma_{ECE}<1\%$. Results for each domain split on DomainNet are included in the appendix, where we observe that $\sigma_{ECE}<0.4\%$. We see a much larger $\sigma_{ECE}$ across domain splits in Table~\ref{tab:officehome-ece-average} and Table~\ref{tab:DomainNet-ece}, which measures the combined variations from samples and domains. It is clear that domain variations dominate the variance of the ECE scores. 

To evaluate the effectiveness of calibration transfer, we define improvement ratio (IR) as  
\begin{equation}
IR=\frac{ECE_{S} - ECE}{ECE_S -ECE_T},
\zerodisplayskips
\end{equation}\label{eq:IR}
where $ECE_S$ and $ECE_T$ refer to the averaged ECE scores obtained via source-only calibration and target-only calibration, respectively. Without calibration transfer, we start from the performance of source-only calibration. Using calibration transfer, we want to approach the performance of target-only calibration. The IR metric evaluates where the performance of a calibration transfer method is located with respect to these starting and ending points. Table~\ref{tab:ir} lists the IRs evaluated on Office-Home and DomainNet.\footnote{Additional results including alternative metrics and confidence intervals can be found in the appendix.}

\noindent{\bf Comparison against source-only calibration.} Without calibration transfer, directly using the temperature learned from the source distribution fails for both datasets. It can even potentially lead to larger errors in comparison to uncalibrated models (comparing the first and second rows in Table~\ref{tab:officehome-ece-average}). 
Compared to uncalibrated and source-only, our algorithms can improve the generalization performance of calibration and achieve lower ECEs for both Office-Home (a reduction of 3.92 percentage points in ECE from the source-only calibration in Table~\ref{tab:officehome-ece-average}) and DomainNet (a reduction of 10.27 percentage points in ECE from the source-only calibration in Table~\ref{tab:DomainNet-ece}).

\begin{table}[t]
\small
\setlength{\tabcolsep}{2pt}
\begin{center}
\begin{tabular}{l|ccc|ccc}
\hline
$\times\sigma_{ECE}$&$<-3$&$(-3,-2)$&$(-2,0)$&$(0,2)$&$(2,3)$&$>3$\\
\hline
TransCal~\cite{wang20}&1&0&0&1&0&10\\
WTS~\cite{pampari20}&1&1&0&1&1&8\\
\hline
\end{tabular}
\vspace{-1em}
\end{center}
\vspace{-1em}
\caption{Number of Office-home domain splits with reduction in ECE achieved by our methods, with $95\%$ confidence ($2\sigma_{ECE}$) and $99\%$ confidence ($3\sigma_{ECE}$).}
\label{tab:officehome-comp}
\vspace{-2em}
\end{table}

\noindent{\bf Comparison against domain adaptation methods.} For Office-Home, we compare the performance of our methods against two recent calibration transfer methods designed for domain adaptation: TransCal~\cite{wang20} and Weighted Temperature Scaling 
(WTS)~\cite{pampari20}. 
On average, we achieve a reduction of $8.86$ percentage points in ECE against TransCal and a reduction of $3.25$ percentage points in ECE against WTS (Table~\ref{tab:officehome-ece-average}). 
While $\sigma_{ECE}$ across domain splits is relatively high, the reduction in ECE achieved by our algorithms is still significant, considering that $\sigma_{ECE}<1\%$ for each domain split. Comparing to TransCal, the reduction in ECE is larger than $3\sigma_{ECE}$ for 10 out of 12 domain splits (i.e., with a confidence $>99\%$ for $83\%$ tested cases in Table~\ref{tab:officehome-comp}). For WTS, the reduction in ECE is larger than $2\sigma_{ECE}$ for 9 out of the 12 domain splits (i.e., with a confidence $>95\%$ for $75\%$ tested cases in 
Table~\ref{tab:officehome-comp}). 

\begin{table}[t]
\small
\setlength{\tabcolsep}{2pt}
\begin{center}
\begin{tabular}{l|r|r}
\hline
&Office-Home&DomainNet\\
\hline
TransCal~\cite{wang20}&0.25&-\\
WTS~\cite{pampari20}&-0.03&-\\
\hline
Set-level&0.56&0.61\\
Cluster-level NN&0.53&0.70\\
Cluster-level Regression&0.54&0.51\\
\hline
Ensemble& 0.60 & 0.69\\
\hline
\end{tabular}
\vspace{-1em}
\end{center}
\vspace{-1em}
\caption{Improvement ratio based on averaged ECE scores.} 
\label{tab:ir}
\vspace{-2em}
\end{table}

As for the IRs in Table~\ref{tab:ir}, TransCal is able to compensate for about one fourth (IR=$25\%$) of the difference between the ECE scores of the target-only and source-only calibration, whereas WTS turns out to perform only comparably with the source-only calibration (IR=$-3\%$)\footnote{As TransCal and WTS reported different ECE scores for source-only and target-only calibration, we use their respective numbers to compute the improvement ratios for fair comparison.}. As expected, our methods yield a higher IR (IR=$60\%$, an improvement of $35$ percentage points over TransCal), compensating for slightly more than half of the differences between the ECE scores of the source-only and the target-only calibration. 

\noindent{\bf Comparison among our methods. }As expected, our ensemble method performs the best (on Office-Home) or on par with the best performance (on DomainNet). Comparing set-level and cluster-level methods, they achieve better performance with respect to different domain splits. On Office-Home (Table~\ref{tab:officehome-ece-average}), the set-level method achieves an averaged ECE of $6.80\%$ and IR of $56\%$ whereas the cluster-level NN method yields an averaged ECE of $6.98\%$ and IR of $53\%$. 

On DomainNet, the cluster-level NN method achieves better performance over several different target domains (Table~\ref{tab:DomainNet-ece}). It produces an averaged ECE of $6.44\%$ and IR of $70\%$ (compensating for $70\%$ of the calibration errors caused by domain shifts). This verifies that the strategy of learning multiple calibration models at the cluster level and using the nearest neighbor algorithm to select the most proper temperature for each test sample can effectively improve calibration performance. 

The cluster-level regression-based method produces slightly higher errors than the other two methods on average. Conceptually, learning a regression model that captures the underlying mapping from features to corresponding optimal temperatures can allow run-time extrapolation such that, instead of selecting from temperatures learned for clusters, one can directly predict a proper temperature for the specific test instance. In practice, its performance is more sensitive to the accuracy and robustness of the learned regression model, clusters and features used. 
More parameters also need to be estimated, compared to temperature scaling that estimates only a single parameter. 

\noindent{\bf Comparison across different domains.}
In Table~\ref{tab:officehome-ece-average} and Table~\ref{tab:DomainNet-ece}, domains are arranged from left to right with increasing image realism. For both datasets, the lowest calibration errors are achieved for domains that reside in the middle of the spectrum. For example, Art has the lowest ECE on Office-Home, whereas Sketch has the lowest ECE on DomainNet. These observations agree with our theoretical analysis, which states that the overlap of the data distributions between the target and calibration domains determines the calibration error. For domains at the ends of the spectrum, chances of obtaining good alignment using the remaining domains decrease. This directly leads to the observed U shape of ECE scores across the domain spectrum (i.e., low in the middle and high at both ends). 

In comparison to domain adaptation, we assume the availability of multiple source domains. Our assumption is more realistic for applications such as recognition using new sensor platforms or autonomous driving under extreme weather/unexplored terrains. 
If unlabeled target data is available at the calibration stage, we can use it to estimate the density ratio. In this setting, our method reduces to calibration transfer via domain adaptation but with the capability to choose the right calibration domain or portion of it to optimize the transfer.
It is also worth noting that we bypass the step of optimizing feature alignment, a commonly used method for domain generalization. Instead, we focus on improving the confidence prediction to better match the classification accuracy, given any classifiers whether or not optimized to maintain accuracy across domains. 
Our calibration methods can be applied on top of a feature space that is aligned across domains to further reduce the remaining misalignment from the calibration point of view. We include additional experiments and discussion in the appendix.

\vspace{-3mm}
\section{Conclusions}
\vspace{-2mm}
In this work, we addressed the problem of confidence calibration for domain generalization, a more challenging problem than calibration for domain adaption as no data from the target domain is used. Our key idea is to exploit multiple calibration domains with covariate shifts against the source domain used for training the classification model and between each other. We compared the proposed solutions under the same theoretical framework against calibration methods based on domain adaptation. We showed that introducing multiple calibration domains can effectively reduce the variance of the density ratio, the main factor that determines the upper bound of the calibration error against the oracle.  
Encouraged by our theoretical study, we proposed three alternative algorithms based on temperature scaling, namely set-level, cluster-level with nearest neighbor, and cluster-level with linear regression. Through experiments using the Office-Home and DomainNet datasets, we demonstrated that our methods can outperform calibration methods via domain adaptation with statistically significant (with a confidence $>95\%$) improvement for at least $75\%$ of the tested scenarios. 

\vspace{-2mm}
\section{Acknowledgments}
\vspace{-2mm}
This material is based upon work supported by the Defense Advanced Research Projects Agency (DARPA) under Contract No. HR001119C0112. Any opinions, findings and conclusions or recommendations expressed in this material are those of the author(s) and do not necessarily reflect the views of the DARPA.
{\small
\bibliographystyle{ieee_fullname}
\bibliography{egbib}

\begin{thebibliography}{10}\itemsep=-1pt

\bibitem{alexandari20}
Amr Alexandari, Anshul Kundaje, and Avanti Shrikumarn.
\newblock Maximum likelihood with bias-corrected calibration is hard-to-beat at
  label shift adaptation.
\newblock In {\em International Conference on Machine Learning}, 2020.

\bibitem{amodei16}
Dario Amodei, Chris Olah, Jacob Steinhardt, Paul Christiano, John Schulman, and
  Dan Mané.
\newblock Concrete problems in ai safety.
\newblock In {\em arXiv:1606.06565}, 2016.

\bibitem{balaji18}
Yogesh Balaji, Swami Sankaranarayanan, and Rama Chellappa.
\newblock Metareg: Towards domain generalization using meta-regularization.
\newblock In {\em Adv. Neural Inform. Process. Syst.}, 2018.

\bibitem{cortes10}
Corinna Cortes, Yishay Mansour, and Mehryar Mohri.
\newblock Learning bounds for importance weighting.
\newblock In {\em Adv. Neural Inform. Process. Syst.}, 2010.

\bibitem{csurka17}
Gabriela Csurka.
\newblock Domain adaptation for visual applications: A comprehensive survey.
\newblock In {\em Domain Adaptation in Computer Vision Applications}, 2017.

\bibitem{dou19}
Qi Dou, Daniel~C. Castro, Konstantinos Kamnitsas, and Ben Glocker.
\newblock Domain generalization via model-agnostic learning of semantic
  features.
\newblock In {\em Adv. Neural Inform. Process. Syst.}, 2019.

\bibitem{finn17}
Chelsea Finn, Pieter Abbeel, and Sergey Levine.
\newblock Model-agnostic meta-learning for fast adaptation of deep networks.
\newblock In {\em International Conference on Machine Learning}, 2017.

\bibitem{ghifary15}
Muhammad Ghifary, W.~Bastiaan Kleijn, Mengjie Zhang, and David Balduzzi.
\newblock Domain generalization for object recognition with multi-task
  autoencoders.
\newblock In {\em Int. Conf. Comput. Vis.}, 2015.

\bibitem{guo17}
Chuan Guo, Geoff Pleiss, Yu Sun, and Kilian~Q. Weinberger.
\newblock On calibration of modern neural networks.
\newblock In {\em International Conference on Machine Learning}, 2017.

\bibitem{resnet}
Kaiming He, Xiangyu Zhang, Shaoqing Ren, and Jian Sun.
\newblock Deep residual learning for image recognition.
\newblock In {\em CVPR}, 2016.

\bibitem{hoffman18}
Judy Hoffman, Eric Tzeng, Taesung Park, Jun-Yan Zhu, Phillip Isola, Kate
  Saenko, Alexei Efros, and Trevor Darrell.
\newblock Cycada: Cycle-consistent adversarial domain adaptation.
\newblock In {\em International Conference on Machine Learning}, 2018.

\bibitem{bhavya19}
Bhavya Kailkhura, Brian Gallagher, Sookyung Kim, Anna Hiszpanski, and
  T.~Yong-Jin Han.
\newblock Reliable and explainable machine-learning methods for accelerated
  material discovery.
\newblock In {\em npj Comput. Mater.}, 2019.

\bibitem{kang19}
Guoliang Kang, Lu Jiang, Yi Yang, and Alexander~G. Hauptmanng.
\newblock Contrastive adaptation network for unsupervised domain adaptation.
\newblock In {\em IEEE Conf. Comput. Vis. Pattern Recog.}, 2019.

\bibitem{mldg}
Da Li, Yongxin Yang, Yi-Zhe Song, and Timothy~M. Hospedales.
\newblock Learning to generalize: Meta-learning for domain generalization.
\newblock In {\em AAAI}, 2018.

\bibitem{li18}
Haoliang Li, Sinno~Jialin Pan, Shiqi Wang, and Alex~C. Kot.
\newblock Domain generalization with adversarial feature learning.
\newblock In {\em IEEE Conf. Comput. Vis. Pattern Recog.}, 2018.

\bibitem{kmeans}
Stuart~P Lloyd.
\newblock Least squares quantization in {PCM}.
\newblock In {\em IEEE Transactions on Information Theory}, 1982.

\bibitem{long18}
Mingsheng Long, Zhangjie Cao, Jianmin Wang, and Michael~I. Jordan.
\newblock Conditional adversarial domain adaptation.
\newblock In {\em Adv. Neural Inform. Process. Syst.}, 2018.

\bibitem{long17}
Mingsheng Long, Han Zhu, Jianmin Wang, and Michael~I. Jordan.
\newblock Deep transfer learning with joint adaptation networks.
\newblock In {\em International Conference on Machine Learning}, 2017.

\bibitem{ECE}
Mahdi~Pakdaman Naeini, Gregory~F Cooper, and Milos Hauskrecht.
\newblock Obtaining well calibrated probabilities using {B}ayesian binning.
\newblock In {\em AAAI}, 2015.

\bibitem{pampari20}
Anusri Pampari and Stefano Ermon.
\newblock Unsupervised calibration under covariate shift.
\newblock In {\em arXiv:2006.16405}, 2020.

\bibitem{pan19}
Yingwei Pan, Ting Yao, Yehao Li, Yu Wang, Chong-Wah Ngo, and Tao Mei.
\newblock Transferrable prototypical networks for unsupervised domain
  adaptation.
\newblock In {\em IEEE Conf. Comput. Vis. Pattern Recog.}, 2019.

\bibitem{park20}
Sangdon Park, Osbert Bastani, James Weimer, and Insup Leen.
\newblock Calibrated prediction with covariate shift via unsupervised domain
  adaptation.
\newblock In {\em AISTATS}, 2020.

\bibitem{DomainNet}
Xingchao Peng, Qinxun Bai, Xide Xia, Zijun Huang, Kate Saenko, and Bo Wang.
\newblock Moment matching for multi-source domain adaptation.
\newblock In {\em Int. Conf. Comput. Vis.}, 2019.

\bibitem{platt99}
John~C. Platt.
\newblock Probabilistic outputs for support vector machines and comparisons to
  regularized likelihood methods.
\newblock In {\em Advances in Large Margin Classifiers}, 1999.

\bibitem{renyi60}
Alfréd Rényi.
\newblock On measures of information and entropy.
\newblock In {\em Proceedings of the 4th Berkeley Symposium on Mathematics,
  Statistics and Probability}, 1960.

\bibitem{shriva17}
Ashish Shrivastava, Tomas Pfister, Oncel Tuzel, Josh Susskind, Wenda Wang, and
  Russ Webb.
\newblock Learning from simulated and unsupervised images through adversarial
  training.
\newblock In {\em IEEE Conf. Comput. Vis. Pattern Recog.}, 2017.

\bibitem{sun19}
Yu Sun, Eric Tzeng, Trevor Darrell, and Alexei~A. Efros.
\newblock Unsupervised domain adaptation through self-supervision.
\newblock In {\em arXiv:1909.11825}, 2019.

\bibitem{toreini20}
Ehsan Toreini, Mhairi Aitken, Kovila P.~L. Coopamootoo, Karen Elliott,
  Vladimiro~Gonzalez Zelaya, Paolo Missier, Magdalene Ng, and Aad van Moorsel.
\newblock Technologies for trustworthy machine learning: A survey in a
  socio-technical context.
\newblock In {\em arXiv:2007.08911}, 2020.

\bibitem{touvron19}
Hugo Touvron, Andrea Vedaldi, Matthijs Douze, and Hervé Jégou.
\newblock Fixing the train-test resolution discrepancy.
\newblock In {\em Adv. Neural Inform. Process. Syst.}, 2019.

\bibitem{officehome}
Hemanth Venkateswara, Jose Eusebio, Shayok Chakraborty, and Sethuraman
  Panchanathan.
\newblock Deep hashing network for unsupervised domain adaptation.
\newblock In {\em IEEE Conf. Comput. Vis. Pattern Recog.}, 2017.

\bibitem{sc2}
Oriol Vinyals, Timo Ewalds, Sergey Bartunov, Petko Georgiev, Alexander~Sasha
  Vezhnevets, Michelle Yeo, Alireza Makhzani, Heinrich Küttler, John Agapiou,
  Julian Schrittwieser, John Quan, Stephen Gaffney, Stig Petersen, Karen
  Simonyan, Tom Schaul, Hado van Hasselt, David Silver, Timothy Lillicrap,
  Kevin Calderone, Paul Keet, Anthony Brunasso, David Lawrence, Anders Ekermo,
  Jacob Repp, and Rodney Tsing.
\newblock Starcraft {II}: A new challenge for reinforcement learning.
\newblock In {\em arXiv, 1708.04782}, 2017.

\bibitem{wang18}
Mei Wang and Weihong Deng.
\newblock Deep visual domain adaptation: A survey.
\newblock In {\em Neurocomputing}, 2018.

\bibitem{wang20}
Ximei Wang, Mingsheng Long, Jianmin Wang, and Michael~I. Jordan.
\newblock Transferable calibration with lower bias and variance in domain
  adaptation.
\newblock In {\em NeurIPS}, 2020.

\bibitem{wilson20}
Garrett Wilson and Diane~J. Cook.
\newblock A survey of unsupervised deep domain adaptation.
\newblock In {\em ACM Transactions on Intelligent Systems and Technology},
  2020.

\bibitem{xie20}
Qizhe Xie, Minh-Thang Luong, Eduard Hovy, and Quoc~V. Le1.
\newblock Self-training with noisy student improves imagenet classification.
\newblock In {\em IEEE Conf. Comput. Vis. Pattern Recog.}, 2020.

\bibitem{zad01}
Bianca Zadrozny and Charles Elkan.
\newblock Obtaining calibrated probability estimates from decision trees and
  naive {B}ayesian classifiers.
\newblock In {\em International Conference on Machine Learning}, 2001.

\bibitem{zad02}
Bianca Zadrozny and Charles Elkan.
\newblock Transforming classifier scores into accurate multiclass probability
  estimates.
\newblock In {\em KDD}, 2002.

\bibitem{zhang20}
Jize Zhang, Bhavya Kailkhura, and T.~Yong-Jin Han.
\newblock Mix-n-match : Ensemble and compositional methods for uncertainty
  calibration in deep learning.
\newblock In {\em International Conference on Machine Learning}, 2020.

\end{thebibliography}
}
\onecolumn
\newpage
\twocolumn
\appendix
\section{Appendix}
\setcounter{table}{0}
\renewcommand\thetable{A.\arabic{table}}

\subsection{Calibration of improved classifiers}
In this section, we provide additional results and discussion on applying proposed calibration algorithms on improved classifiers trained to maintain accuracy across domains.
\subsubsection{Domain generalization accuracy}

In Section~\ref{sec:results} we report experiments with classifiers not specifically optimized with respect to domain generalization accuracy.
Table~\ref{officehome-accuracy} lists corresponding classification accuracy on Office-Home~\cite{officehome}. As we assume that no target
data is available at the training and calibration stage, the
classifier is not adapted to the target domain. As a result,
classification accuracy is relatively low compared to the
reported performance of TransCal~\cite{wang20} where unlabeled target
data is used for unsupervised domain adaptation.

\begin{table}[!htb]
\centering
\small
\vspace{-1em}
\begin{tabular}{cccc|c}
\hline
 Real & Product & Art & Clipart & \textbf{Average} \\
\hline
 62.30 & 60.14 & 45.78 & 37.97 & 51.55 \\
\hline
\end{tabular}
\caption{Domain generalization accuracy (\%) on Office-Home~\cite{officehome}}.
\label{officehome-accuracy}
\vspace{-2em}
\end{table}

\subsubsection{multi-source classifiers}
In addition to experimental set-ups described in Section~\ref{sec:results}, we study the alternative set-up considering multiple source domains. For experiments on Office-Home~\cite{officehome}, each classifier is trained and calibrated on 3 domains and tested on the holdout target domain. The classification accuracy and ECE scores are listed in Table~\ref{tab:multi-source}. Comparing Table~\ref{officehome-accuracy} and ~\ref{tab:multi-source}, we see that classifiers trained on multiple sources produce better generalization accuracy. They also have slightly lower ECE on Real and Product, which shows that the calibration performance of classifiers trained using multiple sources can, to some extent, generalize to new domains, especially for domains that are similar to their training data (ResNet is pre-trained on ImageNet). However, for domains that are more different (Art, Clipart), applying the proposed calibration methods to classifiers trained using a single domain produces lower ECE. 

\begin{table}[!htb]
\centering
\small
\vspace{-1em}
\begin{tabular}{l|rrrr|r}
\hline
 & Real & Product & Art & Clipart & \textbf{Average} \\
\hline
Accuracy & 75.55 & 73.66 & 59.17 & 45.03 & 63.35\\\hline
ECE & 3.35 & 2.80 & 5.64 & 16.01 & 6.95\\
\hline
\end{tabular}
\caption{Calibration of multi-source classifiers on Office-Home~\cite{officehome}}.
\label{tab:multi-source}
\vspace{-2.5em}
\end{table}

\subsubsection{classifiers with domain-invariant features}
We further study the effect of applying proposed calibration algorithms on top of domain-invariant features learned via domain generalization~\cite{mldg}. To better manifest the benefit from both domain-invariant features and calibration, we collect a dataset with 10 domains from simulations in the game environment StarCraft2~\cite{sc2}. For each domain, we consider a different StarCraft2 map and collect data over 6 different unit formation classes. For each experiment we use 4 domains for training the classifiers, 4 domains for calibrating the classifiers and 2 holdout domains for testing. We learn MLP classifiers on top of ResNet18 pretrained on ILSVRC-1000. Table~\ref{tab:improved-features} compares the calibration performance with and without domain-invariant features. It is shown that with domain-invariant features, ECE is further reduced, which indicates that calibration and feature alignment can be complementary. 

We also note that, in general, it is a design choice on how to distribute domains between training and calibration. In this paper, we
focus on one end of the trade-off, i.e., calibration.

\begin{table}[!htb]
\centering
\small
\vspace{-1em}
\begin{tabular}{l|rr|r}
\hline
 ECE (\%)& Uncalibrated & Target-Only & Ours\\ 
\hline
w/o & 23.36 & 5.17 & 6.30\\
\hline
w & 11.63 & 2.96 & 4.41 \\
\hline
\end{tabular}
\caption{Performance comparison on StarCraft2. Four domains for training, four for calibration, and two as the target domains.}.
\label{tab:improved-features}
\vspace{-2.5em}
\end{table}

\onecolumn

\begin{center}
\begin{tabular}{r|rrr|rrr}
\hline
\shortstack{Source\\$\rightarrow$\text{Target}}& uncalibrated & source-only & \shortstack{target-only\\(oracle)} & \textbf{Set-level} & \shortstack{\textbf{Cluster}\\\textbf{-level:NN}} & \shortstack{\textbf{Cluster-level}\\\textbf{Regression}} \\
\hline
\text{I, S}$\rightarrow$ \text{Q} & 21.33$\pm$0.20 & 19.00$\pm$0.20& 0.28$\pm$0.13 & 10.81$\pm$0.17* & \textbf{10.60}$\pm$0.17* & 19.88$\pm$0.20\\
\text{I, C}$\rightarrow$ \text{Q} & 27.12$\pm$0.19 & 23.46$\pm$0.18& 0.53$\pm$0.13 & 11.08$\pm$0.14* & \textbf{6.46}$\pm$0.14* & 9.21$\pm$0.14* \\
\text{I, P}$\rightarrow$ \text{Q} & 26.49$\pm$0.17 & 23.22$\pm$0.16& 0.99$\pm$0.10 & 14.35$\pm$0.13* & \textbf{8.03}$\pm$0.11* & 8.79$\pm$0.11* \\
\text{I, R}$\rightarrow$ \text{Q} & 23.41$\pm$0.18 & 23.67$\pm$0.19& 0.55$\pm$0.12 & 12.34$\pm$0.14* & \textbf{6.24}$\pm$0.12* & 7.68$\pm$0.13* \\
\text{S, C}$\rightarrow$ \text{Q} & 19.56$\pm$0.21 & 19.32$\pm$0.21& 0.97$\pm$0.17 & \textbf{6.92}$\pm$0.18* & 8.78$\pm$0.17* & 16.98$\pm$0.19* \\
\text{S, P}$\rightarrow$ \text{Q} & 20.95$\pm$0.19 & 18.46$\pm$0.18& 0.34$\pm$0.14 & \textbf{10.36}$\pm$0.16* & 11.47$\pm$0.16* & 22.86$\pm$0.20 \\
\text{S, R}$\rightarrow$ \text{Q} & 15.59$\pm$0.18 & 15.66$\pm$0.18& 0.81$\pm$0.15 & \textbf{7.17}$\pm$0.16* & 10.22$\pm$0.17* & 12.29$\pm$0.20* \\
\text{C, P}$\rightarrow$ \text{Q} & 19.56$\pm$0.16 & 18.03$\pm$0.16& 0.69$\pm$0.14 & 9.60$\pm$0.14* & \textbf{5.90}$\pm$0.14* & 8.50$\pm$0.15* \\
\text{C, R}$\rightarrow$ \text{Q} & 18.77$\pm$0.17 & 19.54$\pm$0.17& 0.67$\pm$0.14 & 7.48$\pm$0.14* & 6.74$\pm$0.14* &\textbf{5.87}$\pm$0.14*\\
\text{P, R}$\rightarrow$ \text{Q} & 21.37$\pm$0.16 & 22.01$\pm$0.17& 0.96$\pm$0.11 & 9.95$\pm$0.12* & 6.57$\pm$0.12* & \textbf{5.15}$\pm$0.11* \\
\hline
\text{Q, S}$\rightarrow$ \text{I} & 28.73$\pm$0.15 & 28.73$\pm$0.15& 2.20$\pm$0.12 & 10.73$\pm$0.13* & \textbf{8.07}$\pm$0.13* & 9.71$\pm$0.14* \\
\text{Q, C}$\rightarrow$ \text{I} & 23.96$\pm$0.16 & 26.94$\pm$0.17& 1.41$\pm$0.14 & 8.95$\pm$0.14* & \textbf{6.45}$\pm$0.14* & 7.46$\pm$0.14* \\
\text{Q, P}$\rightarrow$ \text{I} & 20.07$\pm$0.14 & 21.77$\pm$0.15& 1.93$\pm$0.13 & 8.34$\pm$0.16* & \textbf{5.05}$\pm$0.13* & 5.61$\pm$0.14* \\
\text{Q, R}$\rightarrow$ \text{I} & 25.94$\pm$0.17 & 30.28$\pm$0.18& 1.40$\pm$0.14 & 11.98$\pm$0.16* & 6.39$\pm$0.15* & \textbf{3.50}$\pm$0.14* \\
\text{S, C}$\rightarrow$ \text{I} & 32.63$\pm$0.19 & 32.33$\pm$0.19& 1.82$\pm$0.15 & \textbf{11.55}$\pm$0.16* & 13.14$\pm$0.16* & 12.10$\pm$0.17* \\
\text{S, P}$\rightarrow$ \text{I} & 23.01$\pm$0.17 & 20.15$\pm$0.17& 1.82$\pm$0.14  & \textbf{6.79}$\pm$0.15* & 9.62$\pm$0.15* & 12.55$\pm$0.16* \\
\text{S, R}$\rightarrow$ \text{I} & 19.36$\pm$0.16 & 19.45$\pm$0.16& 2.35$\pm$0.15 & \textbf{3.97}$\pm$0.15* & 9.72$\pm$0.16* & 27.63$\pm$0.22 \\
\text{C, P}$\rightarrow$ \text{I} & 20.87$\pm$0.18 & 19.12$\pm$0.16& 1.85$\pm$0.14 & \textbf{5.61}$\pm$0.16* & 6.97$\pm$0.16* & 10.92$\pm$0.17* \\
\text{C, R}$\rightarrow$ \text{I} & 23.88$\pm$0.18 & 24.82$\pm$0.18& 1.50$\pm$0.13 & \textbf{4.07}$\pm$0.16* & 7.39$\pm$0.17* & 9.78$\pm$0.18* \\
\text{P, R}$\rightarrow$ \text{I} & 21.48$\pm$0.18 & 22.24$\pm$0.18& 1.85$\pm$0.14 & \textbf{1.93}$\pm$0.15* & 6.31$\pm$0.16* & 20.05$\pm$0.22* \\
\hline
\text{Q, I}$\rightarrow$ \text{S} & 15.21$\pm$0.22 & 17.13$\pm$0.22& 0.85$\pm$0.16 & 4.33$\pm$0.21* & 2.65$\pm$0.20* & \textbf{2.06}$\pm$0.19* \\
\text{Q, C}$\rightarrow$ \text{S} & 17.66$\pm$0.24 & 20.56$\pm$0.24& 1.96$\pm$0.22 & 1.63$\pm$0.21* & \textbf{1.04}$\pm$0.19* & 1.28$\pm$0.20* \\
\text{Q, P}$\rightarrow$ \text{S} & 18.57$\pm$0.23 & 20.39$\pm$0.24& 1.35$\pm$0.20 & 5.38$\pm$0.22* & 3.32$\pm$0.23* & \textbf{3.10}$\pm$0.22* \\
\text{Q, R}$\rightarrow$ \text{S} & 15.03$\pm$0.23 & 19.03$\pm$0.23& 2.78$\pm$0.21 & 1.45$\pm$0.20* & \textbf{1.31}$\pm$0.19* & 2.72$\pm$0.22* \\
\text{I, C}$\rightarrow$ \text{S} & 20.89$\pm$0.25 & 17.43$\pm$0.25& 1.47$\pm$0.22 & \textbf{1.14}$\pm$0.20* & 3.10$\pm$0.24* & 6.78$\pm$0.25* \\
\text{I, P}$\rightarrow$ \text{S} & 18.03$\pm$0.24 & 14.83$\pm$0.24& 1.48$\pm$0.22 & \textbf{0.93}$\pm$0.17* & 4.75$\pm$0.22* & 6.67$\pm$0.24* \\
\text{I, R}$\rightarrow$ \text{S} & 16.33$\pm$0.23 & 16.59$\pm$0.23& 2.32$\pm$0.22 & \textbf{3.27}$\pm$0.23* & 4.14$\pm$0.23* & 16.69$\pm$0.28 \\
\text{C, P}$\rightarrow$ \text{S} & 16.47$\pm$0.25 & 14.59$\pm$0.25& 2.09$\pm$0.23 & \textbf{1.43}$\pm$0.21* & 3.31$\pm$0.20* & 7.84$\pm$0.25* \\
\text{C, R}$\rightarrow$ \text{S} & 14.07$\pm$0.24 & 14.94$\pm$0.24& 3.97$\pm$0.23 & 7.23$\pm$0.23* & \textbf{4.97}$\pm$0.24* & 10.12$\pm$0.27* \\
\text{P, R}$\rightarrow$ \text{S} & 14.27$\pm$0.23 & 14.98$\pm$0.23& 1.83$\pm$0.22 & 8.36$\pm$0.23* & \textbf{1.82}$\pm$0.20* & 15.65$\pm$0.29 \\
\end{tabular}
\captionof{table}{Quantitative evaluation of calibration on the DomainNet dataset~\cite{DomainNet} for experiments using Quickdraw (Q), Infograph (I) or Sketch (S) as the target domain. For each experiment, we evaluate each of the three proposed algorithms over 1000 experiments with randomly selected test data from the target domain. Results with statistically significant improvement against source-only method are highlighted with asterisks.}
\end{center}

\begin{table*}[t!]
\centering
\begin{tabular}{r|rrr|rrr}
\hline
\shortstack{Source\\$\rightarrow$\text{Target}} & uncalibrated & source-only & \shortstack{target-only\\(oracle)} & \textbf{Set-level} & \shortstack{\textbf{Cluster}\\\textbf{-level:NN}} & \shortstack{\textbf{Cluster-level}\\\textbf{Regression}} \\
\hline
\text{Q, I}$\rightarrow$ \text{C} & 12.09$\pm$0.17 & 14.05$\pm$0.17& 1.83$\pm$0.16& \textbf{1.00}$\pm$0.14* & 3.23$\pm$0.16* & 4.64$\pm$0.17* \\
\text{Q, S}$\rightarrow$ \text{C} & 17.18$\pm$0.18 & 17.18$\pm$0.18& 2.74$\pm$0.17 & \textbf{2.45}$\pm$0.17* & 3.83$\pm$0.18* & 3.71$\pm$0.17* \\
\text{Q, P}$\rightarrow$ \text{C} & 14.33$\pm$0.18 & 16.09$\pm$0.18& 2.08$\pm$0.18 & \textbf{1.07}$\pm$0.14* & 4.03$\pm$0.18* & 5.08$\pm$0.18* \\
\text{Q, R}$\rightarrow$ \text{C} & 10.05$\pm$0.19 & 13.98$\pm$0.19& 1.84$\pm$0.17 & 5.92$\pm$0.19* & 7.82$\pm$0.19* & \textbf{4.52}$\pm$0.19*\\
\text{I, S}$\rightarrow$ \text{C} & 11.69$\pm$0.20 & 9.21$\pm$0.20& 3.32$\pm$0.19 & 6.01$\pm$0.19* & \textbf{3.94}$\pm$0.19* & 7.51$\pm$0.20*\\
\text{I, P}$\rightarrow$ \text{C} & 12.75$\pm$0.18 & 9.61$\pm$0.18& 2.75$\pm$0.18 & 5.59$\pm$0.18* & \textbf{4.13}$\pm$0.18* & 9.59$\pm$0.20\\
\text{I, R}$\rightarrow$ \text{C} & 10.26$\pm$0.19 & 10.52$\pm$0.19& 1.94$\pm$0.17 & 11.21$\pm$0.19 & \textbf{8.38}$\pm$0.19* & 14.84$\pm$0.21\\
\text{S, P}$\rightarrow$ \text{C} & 11.49$\pm$0.19 & 8.80$\pm$0.19& 2.88$\pm$0.19 & 6.77$\pm$0.19* & \textbf{4.41}$\pm$0.19* & 10.76$\pm$0.20 \\
\text{S, R}$\rightarrow$ \text{C} & 6.74$\pm$0.20 & 6.83$\pm$0.20& 3.37$\pm$0.19 & 12.86$\pm$0.21 & 10.34$\pm$0.20 & 11.57$\pm$0.21 \\
\text{P, R}$\rightarrow$ \text{C} & 9.69$\pm$0.18 & 10.40$\pm$0.18& 2.37$\pm$0.17 & 15.43$\pm$0.20 & 10.52$\pm$0.19 & 12.64$\pm$0.22\\
\hline
\text{Q, I}$\rightarrow$ \text{P} & 15.35$\pm$0.25 & 17.42$\pm$0.25& 1.89$\pm$0.23 &  2.76$\pm$0.24* & 3.42$\pm$0.24* & \textbf{2.03}$\pm$0.23* \\
\text{Q, S}$\rightarrow$ \text{P} & 21.61$\pm$0.27 & 21.60$\pm$0.27& 3.06$\pm$0.25 & 2.24$\pm$0.22* & \textbf{2.21}$\pm$0.22* & 2.60$\pm$0.25*\\
\text{Q, C}$\rightarrow$ \text{P} & 19.89$\pm$0.26 & 22.82$\pm$0.26& 3.01$\pm$0.24 & 3.01$\pm$0.25* & 2.75$\pm$0.25* &\textbf{2.50}$\pm$0.24*\\
\text{Q, R}$\rightarrow$ \text{P} & 12.37$\pm$0.28 & 16.34$\pm$0.28& 2.17$\pm$0.26 & 6.23$\pm$0.28* & \textbf{4.70}$\pm$0.27* & 5.73$\pm$0.28*\\ \text{I,S}$\rightarrow$ \text{P} & 14.93$\pm$0.27 & 12.40$\pm$0.27& 2.88$\pm$0.25 & 3.22$\pm$0.26* & 2.67$\pm$0.26* & \textbf{2.40}$\pm$0.24*\\
\text{I, C}$\rightarrow$ \text{P} & 22.73$\pm$0.28 & 19.30$\pm$0.28& 1.95$\pm$0.24 & \textbf{1.88}$\pm$0.24* & 6.02$\pm$0.27* & 9.58$\pm$0.28* \\
\text{I, R}$\rightarrow$ \text{P} & 12.76$\pm$0.28 & 13.02$\pm$0.28& 3.36$\pm$0.27 & 13.46$\pm$0.29 & \textbf{7.67}$\pm$0.28* & 18.22$\pm$0.32 \\
\text{S, C}$\rightarrow$ \text{P} & 19.35$\pm$0.28 & 19.09$\pm$0.28& 3.39$\pm$0.27 & \textbf{2.34}$\pm$0.25* & 4.77$\pm$0.28* & 3.40$\pm$0.26* \\
\text{S, R}$\rightarrow$ \text{P} & 11.44$\pm$0.28 & 11.53$\pm$0.28& 1.57$\pm$0.23 & 11.06$\pm$0.28 & \textbf{2.12}$\pm$0.25* & 9.35$\pm$0.30*\\
\text{C, R}$\rightarrow$ \text{P} & 13.89$\pm$0.28 & 14.76$\pm$0.28& 3.49$\pm$0.27 & 12.50$\pm$0.28* & \textbf{5.43}$\pm$0.28* & 15.03$\pm$0.32\\
\hline
\text{Q, I}$\rightarrow$ \text{R} & 10.47$\pm$0.36 & 12.56$\pm$0.36& 2.17$\pm$0.31 & 7.49$\pm$0.37* & 8.29$\pm$0.37* &\textbf{4.22}$\pm$0.35*\\
\text{Q, S}$\rightarrow$ \text{R} & 16.68$\pm$0.36 & 16.68$\pm$0.36& 2.51$\pm$0.33 & 8.07$\pm$0.35* & \textbf{6.68}$\pm$0.35* & 7.12$\pm$0.36*\\
\text{Q, C}$\rightarrow$ \text{R} & 13.41$\pm$0.37 & 16.17$\pm$0.37& 2.26$\pm$0.33 & 8.97$\pm$0.36* &10.75$\pm$0.37* & \textbf{6.19}$\pm$0.36*\\
\text{Q, P}$\rightarrow$ \text{R} & 10.78$\pm$0.35 & 12.41$\pm$0.35& 1.98$\pm$0.30 & 10.66$\pm$0.36* & 10.07$\pm$0.38* & \textbf{5.13}$\pm$0.35*\\
\text{I, S}$\rightarrow$ \text{R} & 11.37$\pm$0.38 & 9.06$\pm$0.38& 2.65$\pm$0.33 & 11.36$\pm$0.39 & \textbf{6.32}$\pm$0.36* & 10.92$\pm$0.39\\
\text{I, C}$\rightarrow$ \text{R} & 13.12$\pm$0.38 & 10.11$\pm$0.38& 2.52$\pm$0.33 & 16.17$\pm$0.39 & 12.36$\pm$0.37 & \textbf{7.63}$\pm$0.37*\\
\text{I, P}$\rightarrow$ \text{R} & 10.11$\pm$0.35 & 7.32$\pm$0.35& 2.11$\pm$0.30 & 19.77$\pm$0.37 & 9.15$\pm$0.36 & 24.13$\pm$0.41\\
\text{S, C}$\rightarrow$ \text{R} & 12.52$\pm$0.38 & 12.28$\pm$0.38& 2.30$\pm$0.31 & 15.17$\pm$0.38 & 12.65$\pm$0.37 &\textbf{8.84}$\pm$0.37*\\
\text{S, P}$\rightarrow$ \text{R} & 10.10$\pm$0.35 & 7.72$\pm$0.35& 2.41$\pm$0.29 & 18.23$\pm$0.37 & 4.85$\pm$0.34* & 9.53$\pm$0.38\\
\text{C, P}$\rightarrow$ \text{R} & 9.59$\pm$0.35 & 8.10$\pm$0.35& 2.38$\pm$0.31 & 17.94$\pm$0.37 & 12.34$\pm$0.37 & 11.76$\pm$0.39 \\
\end{tabular}

\caption{Quantitative evaluation of calibration on the DomainNet dataset~\cite{DomainNet} for experiments using Clipart (C), Painting (P) or Real (R) as the target domain. For each experiment, we evaluate each of the three proposed algorithms over 1000 experiments with randomly selected test data from the target domain. Results with statistically significant improvement against source-only method are highlighted with asterisks.}
\end{table*}

\begin{table*}[htbp!]
\hspace{-2em}
\small
\begin{tabular}{r|r|rrr|rrr|rrr|rrr}
\hline
 & & A$\rightarrow$ \text{C} & P$\rightarrow$ \text{C} & R$\rightarrow$ \text{C} & C$\rightarrow$ \text{A} & P$\rightarrow$ \text{A} & R$\rightarrow$ \text{A} & C$\rightarrow$ \text{P} & A$\rightarrow$ \text{P} & R$\rightarrow$ \text{P} & C$\rightarrow$ \text{R} & A$\rightarrow$ \text{R} & P$\rightarrow$ \text{R}\\
\hline
 & Avg. & 11.84 & 15.81 & 16.58 & 7.61 & 12.52 & 7.80 & 5.78 &6.81 & 4.38 & 5.86 & 4.31 & 4.59\\
Uncalibrated & 2.5\% & 10.26 & 14.2 & 14.92 & 6.66 & 11.5 & 6.81 & 4.31& 5.28 & 3.16 & 4.55 & 3.08 & 3.38\\
 & 97.5\% & 13.36 & 17.42 & 18.23 & 8.64 & 13.5 & 8.72 & 7.33 & 8.36 & 5.64 & 7.17 & 5.55 & 5.86\\
 \hline
 & Avg. & 16.95 & 20.3 & 16.82 & 7.37 & 17.05 & 7.96 & 5.57 &10.64 & 4.50 & 5.75 & 5.34 & 7.18\\
Source-only & 2.5\% & 15.39 & 18.7 & 15.17 & 6.41 & 16.00 & 7.00 & 4.08 & 8.94 & 3.28 & 4.48 & 4.03 & 5.84\\
 & 97.5\% & 18.45 & 21.94 & 18.46 & 8.41 & 18.02 & 8.90 & 7.09 & 12.18 & 5.79 & 7.07 & 6.68 & 8.57\\
 \hline
 & Avg. & 4.3 & 4.24 & 3.76 & 4.08 & 3.43 & 3.16 & 3.32 & 5.35 & 3.63 & 3.67 & 4.19 & 4.18\\
Target-only & 2.5\% & 2.98 & 2.99 & 2.51 & 3.13 & 2.56 & 2.39 & 2.25 & 3.89 & 2.51 & 2.51 & 2.97 & 2.86\\
(oracle) & 97.5\% & 5.64 & 5.71 & 5.10 & 5.06 & 4.39 & 3.97 & 4.55 & 6.74 & 4.82 & 4.97 & 5.42 & 5.57\\
\hline
TransCal~\cite{wang20} & - & 22.9 & 40.4 & 4.5 & 21.7 & 18.5 & 21.6 & 14 & 9.3 & 15.6 & 6.4 & 5.1 & 13.9\\ 
WTS~\cite{pampari20} & - & 12.8 & 26.8 & 17.3 & 6.9 & 8.5 & 10.4 & 6.4 & 1.5 & 3.8 & 5.7 & 6.4 & 10.8\\
\hline
& Avg.& 10.98 & 7.71 & 11.19 & 4.50 & 3.68 & 4.44 &3.22 & 5.39 & 8.1 & 3.73 & 7.86 & 10.83\\
\textbf{Set-level} & 2.5\% & 9.39 & 6.24 & 9.56 & 3.68 & 2.82 & 3.53 & 2.15 & 3.99 & 6.74 & 2.57 & 6.37 & 9.27\\
& 97.5\% & 12.49* & 9.29* & 12.81 & 5.48* & 4.64* & 5.41* & 4.45 & 7.03 & 9.65 & 5.06 & 9.49 & 12.29\\
\hline
 & Avg. & 12.54 & 9.12 & 12.64 & 4.72 & 6.92 & 3.05 &3.50 & 6.26 & 5.54 & 4.00 & 6.75 & 8.72\\
\textbf{Cluster-level:} & 2.5\% & 10.96 & 7.56 & 11.06 & 3.81 & 5.91 & 2.22 &2.44 & 4.82 & 4.23 & 2.83 & 5.35 & 7.20\\
\textbf{NN} & 97.5\% & 14 & 10.94* & 14.21 & 5.69* & 7.96* & 3.93* & 4.73 & 7.89 & 6.99 & 5.32 & 8.12 & 10.15\\
\hline
 & Avg. & 13.1 & 10.43 & 12.48 & 4.48 & 5.46 & 3.90 & 3.95 & 6.08 & 5.30 & 3.95 & 6.08 & 8.05\\
\textbf{Cluster-level}: & 2.5\% & 11.55 & 8.85 & 10.84 & \textbf{3.54}& 4.49 & 3.15 & 2.74 & 4.55 & 4.08 & 2.72 & 4.65 & 6.61\\
\textbf{Regression} & 97.5\% & 14.63 & 12.19* & 14.06 & 5.38* & 6.42* & 4.78* & 5.19 & 7.54 & 6.58 & 5.28 & 7.54 & 9.64\\
\hline
& Avg. & 12.53 & 9.10 & 12.28 & 5.02 & 4.36 & 3.04 & 3.25 & 5.57 & 5.27 & 3.82 & 6.25 & 7.88\\ 
\textbf{Ensemble} &2.5\% & 11.01 & 7.56 & 10.69 & 4.15 & 3.44 & \textbf{2.32} & 2.18 & 4.16 & 3.90 & 2.60 & 4.79 & 6.37\\
\textbf{(Avg. logits)} &97.5\% & 14.01 & 10.70 & 13.94 & 5.99 & 5.26 & 3.85* & 4.37 & 7.05 & 6.69 & 5.17 & 7.77 & 9.32\\
\hline

\end{tabular}
\vspace{-1em}
\caption{Confidence intervals of ECE (\%) on  Office-Home~\cite{officehome}. Results with statistically significant improvement against source-only and domain adaptation methods are highlighted with asterisks.
}
\end{table*}

\begin{table*}[t!]
\hspace{-2em}
\small
\begin{tabular}{r|r|rrr|rrr|rrr|rrr}
\hline
& & A$\rightarrow$ \text{C} & P$\rightarrow$ \text{C} & R$\rightarrow$ \text{C} & C$\rightarrow$ \text{A} & P$\rightarrow$ \text{A} & R$\rightarrow$ \text{A} & C$\rightarrow$ \text{P} & A$\rightarrow$ \text{P} & R$\rightarrow$ \text{P} & C$\rightarrow$ \text{R} & A$\rightarrow$ \text{R} & P$\rightarrow$ \text{R}\\
 \hline
 & Avg. & -1.88 & -2.07 & -0.10 & 0.05 & -1.77 & -0.05 & 0.04 &-0.86 & -0.01 & 0.04 & -0.14 & -0.48\\
Source-only & 2.5\% & -2.08 & -2.26 & -0.12 & 0.05 & -1.90 & -0.05 & 0.02 & -1.07 & -0.02 & 0.02 & -0.35 & -0.65\\
 & 97.5\% & -1.67 & -1.88 & -0.09 & 0.06 & -1.64 & -0.04 & 0.05 & -0.63 & -0.01 & 0.05 & -0.07 & -0.33\\
 \hline
 & Avg. & 1.51 & 3.22 & 3.37 & 0.70 & 2.19 & 0.75 & 0.31 & 0.24 & 0.09 & 0.34 & 0.01 & -0.04\\
Target-only & 2.5\% & 0.93 & 2.57 & 2.62 & 0.46 & 1.84 & 0.55 & 0.03 & 0.06 & 0.00 & 0.12 & -0.00 & -0.27\\
(oracle) & 97.5\% & 2.04 & 3.93 & 4.12 & 0.94 & 2.51 & 0.93 & 0.57 & 0.40 & 0.18 & 0.57 & 0.02 & 0.21\\
\hline
& Avg.& 0.25 & 2.55 & 1.89 & 0.61 & 2.19 & 0.51 &0.32 & 0.12 & -0.97 & 0.30 & -0.90 & -1.53\\
\textbf{Set-level} & 2.5\% & 0.21* & 2.22* & 1.64* & 0.49* & 1.83* & 0.19* & 0.07* & -0.19* & -1.44 & 0.02 & -1.23 & -2.03\\
& 97.5\% & 0.28 & 2.91 & 2.13 & 0.73 & 2.52 & 0.80 & 0.57 & 0.42 & -0.50 & 0.59 & -0.59 & -0.96\\
\hline
 & Avg. & -0.21 & 2.41 & 1.55 & 0.58 & 1.03 & 0.68 & 0.59 & 0.08 & -0.77 & 0.46 & -0.63 & -1.21\\
\textbf{Cluster-level:} & 2.5\% & -0.52* & 2.14* & 1.32* & 0.46* & 0.78* & 0.44* & 0.31* & -0.24* & -1.19 & 0.22* & -0.96 & -1.67\\
\textbf{NN} & 97.5\% & 0.09 & 2.73 & 1.78 & 0.72 & 1.25 & 0.90 & 0.86 & 0.41 & -0.37 & 0.72 & -0.33 & -0.73\\
\hline
 & Avg. & -0.39 & 2.30 & 2.15 & 0.82 & 1.16 & 0.84 & 0.80 & 0.05 & -0.65 & 0.38 & -0.70 & -1.70\\
\textbf{Cluster-level:} & 2.5\% & -0.82* & 1.96* & 1.80* & 0.63* & 0.86* & 0.57* & 0.50* & -0.40* & -1.07 & 0.06* & -1.06 & -2.26\\
\textbf{Regression} & 97.5\% & 0.03 & 2.64 & 2.52 & 1.01 & 1.45 & 1.11 & 1.12 & 0.50 & -0.25 & 0.72 & -0.34 & -1.14\\
\hline
& Avg. & 0.05 & 2.52 & 1.94 & 0.76 & 1.74 & 1.02 & 0.72 & 0.34 & -0.45 & 0.47 & -0.47 & -1.04\\
\textbf{Ensemble:} & 2.5\% & -0.18* & 2.24* & 1.71* & 0.63*& 1.48* & 0.79* & 0.47* & 0.02* & -0.84 & 0.22* & -0.74 & -1.50\\
\textbf{(avg. logits)} & 97.5\% & 0.27 & 2.83 & 2.19 & 0.89 & 1.99 & 1.24 & 0.97 & 0.66 & -0.08 & 0.75 & -0.20 & -0.58\\
\hline
\end{tabular}
\vspace{-1em}
\caption{Confidence intervals of calibration gain~\cite{zhang20} (\%) on  Office-Home. Results with statistically significant improvement (higher gain) against source-only method are highlighted with asterisks.
}
\end{table*}

\end{document}